\documentclass[runningheads]{llncs}
\usepackage[T1]{fontenc}
\usepackage{graphicx, color, booktabs,multirow }
\usepackage{dblfloatfix, url}
\usepackage{amssymb,array, dsfont, amsmath, algorithmic, algorithm}
\begin{document}
\title{Learn to Ignore: Domain Adaptation for Multi-Site MRI Analysis}
%
%
\author{Julia Wolleb\inst{1}, 
Robin Sandk\"uhler\inst{1}, 
Florentin Bieder\inst{1}, 
Muhamed Barakovic\inst{1},
Nouchine Hadjikhani\inst{3,4}, 
Athina Papadopoulou\inst{1,2}, 
\"Ozg\"ur Yaldizli\inst{1,2},
Jens Kuhle\inst{2},
Cristina Granziera\inst{1,2},
and Philippe C. Cattin\inst{1}}
\authorrunning{J. Wolleb et al.}
%

\institute{Department of Biomedical Engineering, University of Basel, Allschwil, Switzerland \and University Hospital Basel, Switzerland \and 
  Massachusetts General Hospital, Harvard Medical School, Charlestown MA, USA \and Gillberg Neuropsychiatry Center, Sahlgrenska Academy, University of Gothenburg, Sweden
  \\
\email{julia.wolleb@unibas.ch}
}
\maketitle              
\begin{abstract}
The limited availability of large image datasets, mainly due to data privacy and differences in acquisition protocols or hardware, is a significant issue in the development of accurate and generalizable machine learning methods in medicine. 
This is especially the case for Magnetic Resonance (MR) images, where different MR scanners introduce a bias that limits the performance of a machine learning model.
We present a novel method that learns to ignore the scanner-related features present in MR images, by introducing specific additional constraints on the latent space. We focus on a real-world classification scenario, where only a small dataset provides images of all classes. Our method \textit{Learn to Ignore (L2I)} outperforms state-of-the-art domain adaptation methods on a multi-site MR dataset for a classification task between multiple sclerosis patients and healthy controls.
\end{abstract}

\section{Introduction}
Due to its high soft-tissue contrast, Magnetic Resonance Imaging (MRI) is a powerful diagnostic tool for many neurological disorders. 
However, compared to other imaging modalities like computed tomography, MR images only provide relative values for different tissue types. These relative values depend on the scanner manufacturer, the scan protocol, or even the software version.
We refer to this problem as the scanner bias. While human medical experts can adapt to these relative changes, they represent a major problem for machine learning methods, leading to a low generalization quality of the model.
By defining different scanner settings as different domains, we look at this problem from the perspective of domain adaptation (DA) \cite{bentheory}, where the main task is learned on a \textit{source domain}. The model then should perform well on a different \textit{target domain}.

\subsection{Related Work}\label{RW}
An overview of DA in medical imaging can be found at \cite{review2}. One can generally distinguish between unsupervised domain adaptation (UDA) \cite{ackaouy}, where the target domain data is unlabeled, or supervised domain adaptation (SDA) \cite{sda2}, where the labels of the target domain are used during training.\\
The problem of scanner bias is widely known to disturb the automated analysis of MR images \cite{problem3}, and a lot of work already tackles the problem of multi-site MR harmonization \cite{harm2}.
Deepharmony \cite{deepharmony} uses paired data to change the contrast of MRI from one scanner to another scanner with a modified U-Net. Generative Adversarial Networks aim to generate new images to overcome the domain shift \cite{generateadapt}. These methods modify the intensities of each pixel before training for the main task. This approach is preferably avoided in medical applications, as it bears the risk of removing important pixel-level information required later for other tasks, such as segmentation or anomaly detection.\\
Domain-adversarial neural networks \cite{dann} can be used for multi-site brain lesion segmentation \cite{kamnitsas}. Unlearning the scanner bias \cite{unlearning} is an SDA method for MRI harmonization and improves the performance in age prediction from MR images. 
The introduction of contrastive loss terms \cite{supervised,centerloss,triplet} can also be used for domain generalization \cite{cdd,motiian,dou}. 
Disentangling the latent space has been done for MRI harmonization \cite{tsaftaris,disentangled}. Recently, heterogeneous DA \cite{hda} was also of interest for lesion segmentation \cite{hda_seg}.

\subsection{Problem Statement}\label{problemstate}
All DA methods mentioned in Section \ref{RW} have in common that they must learn the main task on the source domain.
\begin{figure}[b]
	\centering
 \includegraphics[width=\textwidth]{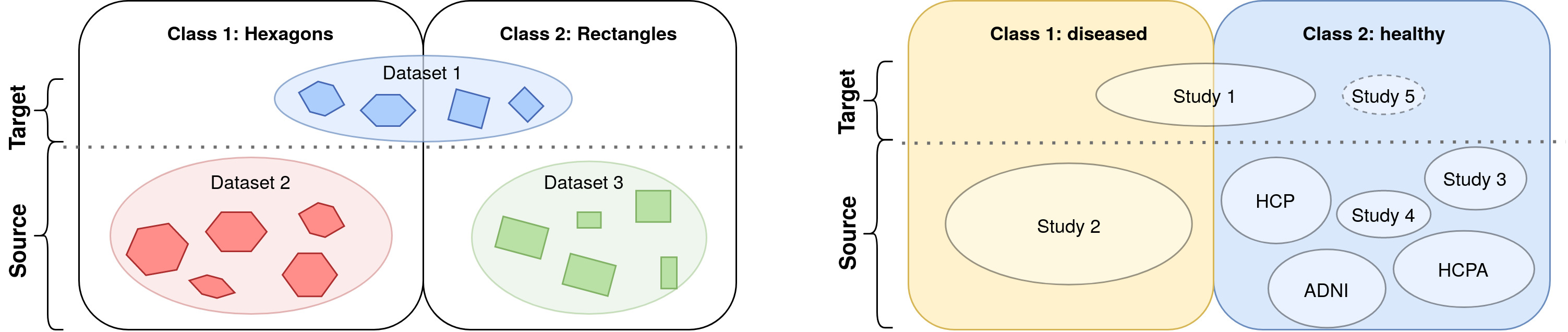}
		\caption{Quantity charts for the datasets in the source and target domain. The chart on the left illustrates the problem, and the chart on the right shows the real-world application on the MS dataset.} 
		  \label{chart}
\end{figure}
 However, it can happen that the bias present in datasets of various origins disturbs the learning of a specific task. $\textrm{Figure \ref{chart}}$ on the left illustrates the problem and the relation of the different datasets on a toy example for the classification task between hexagons and rectangles. Due to the high variability in data, often only a small and specific dataset is at hand to learn the main task: Dataset 1 forms the target domain with only a small number of samples of hexagons and pentagons.  Training on this dataset alone yields a low generalization quality of the model. To increase the number of training samples, we add \mbox{Dataset 2} and \mbox{Dataset 3}. They form the source domain.
 As these additional datasets come from different origins, they differ from each other in color. Note that they only provide either rectangles (Dataset 3) or hexagons (Dataset 2). 
The challenge of such a setup is that during training on the source domain, the color is the dominant feature, and the model learns to distinguish between green and red rather than counting the number of vertices. Classical DA approaches then learn to overcome the domain shift between source and target domain. However, the model will show poor performance on the target domain: The learned features are not helpful, as all hexagons and rectangles are blue in Dataset 1. \\
 This type of problem is highly common in the clinical environment, where different datasets are acquired with different settings, which corresponds to the colors in the toy example. In this project, the main task is to distinguish between multiple sclerosis (MS) patients and healthy controls.
 The quantity chart in $\textrm{Figure \ref{chart}}$ on the right visualizes the different allocations of the MS dataset. Only the small in-house Study 1 provides images of both MS patients and healthy subjects acquired with the same settings. To get more data, we collect images from other in-house studies. As in the hospital mostly data of patients are collected, we add healthy subjects from public datasets, resulting in the presented problem. \\
In this work, we present a new supervised DA method called \textit{Learn to Ignore (L2I)}, which aims to ignore features related to the scanner bias while focusing on disease-related features for a classification task between healthy and diseased subjects. We exploit the fact that the target domain contains images of subjects of both classes with the same origin, and use this dataset to lead the model's attention to task-specific features. We developed specific constraints on the latent space and introduce two novel loss terms that can be added to any classification network.
We evaluate our method on a multi-site MR dataset of MS patients and healthy subjects, compare it to state-of-the-art methods, and perform various ablation studies. The source code is available at \url{https://gitlab.com/cian.unibas.ch/L2I}.

\section{Method}\label{method}
We developed a strategy that aims to ignore features that disturb the learning of a classification task between $n$ classes.
\begin{figure}[h]
	\centerline{\includegraphics[width=1\textwidth]{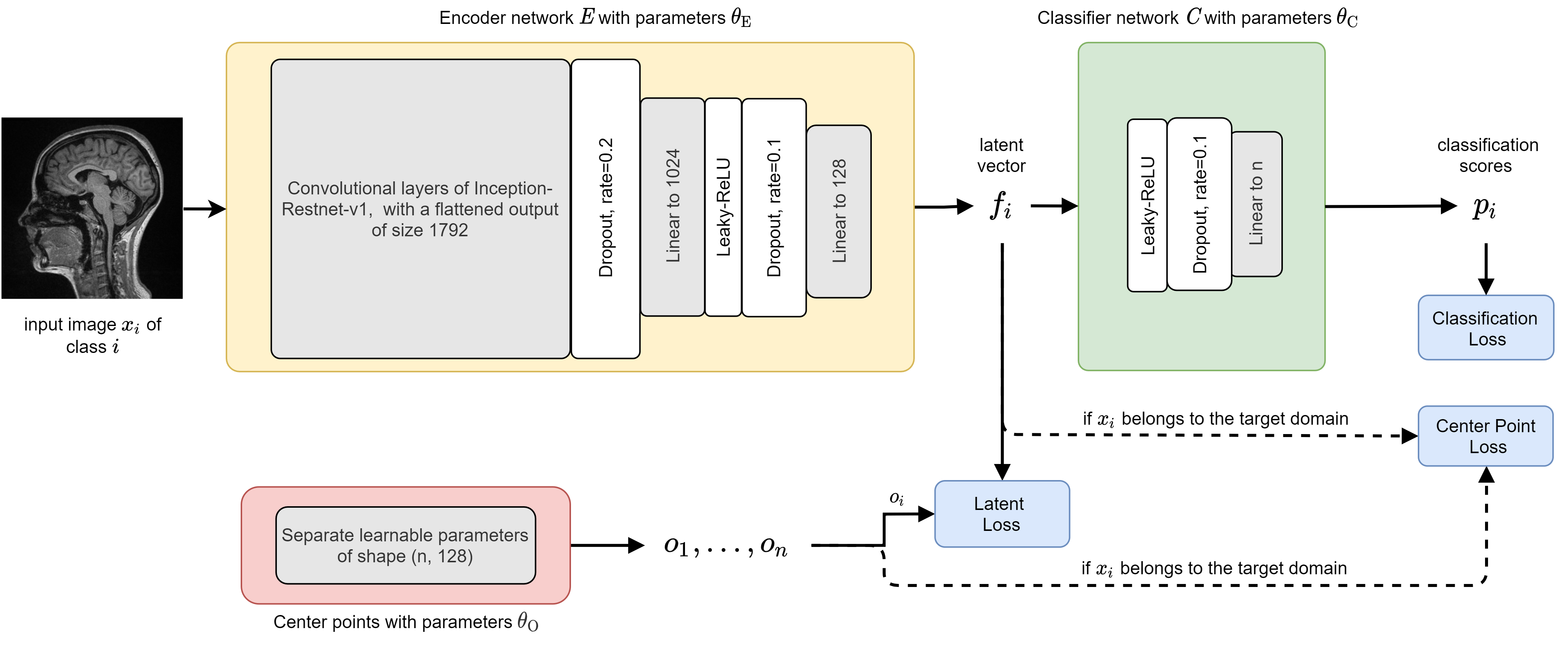}}
	\caption{Architecture of the classification network consisting of an encoder $E$ with parameters $\theta_\textrm{E}$, a fully connected classifier $C$ with parameters $\theta_\textrm{C}$, and separate learnable parameters $\theta_\textrm{O}$. Here, $o_1, ..., o_n$ are learnable center points in the latent space, $f_i$ is a vector in the latent space, and $p_i$ is the classification score for class $i$.  } \label{fig1}
\end{figure}
The building blocks of our setup are shown in Figure \ref{fig1}. 
The input image ${x_i \in \mathbb{R}^3}$  of class ${i \in \{1,...,n\}}$ is the input for the encoder network $E$ with parameters $\theta_\textrm{E}$, which follows the structure of  $\textrm{Inception-ResNet-v1}$ \cite{inception}. However, we replaced the 2D convolutions with 3D convolutions and changed the batch normalization layers to instance normalization layers. The output is the latent vector ${f_i=E(x_i) \in \mathbb{R}^{m}}$, where $m$ denotes the dimension of the latent space. This latent vector is normalized to a length of $1$ and forms the input for the classification network $C$ with parameters $\theta_\textrm{C}$. Finally, we get the classification scores ${p_i=C(f_i)=C(E(x_i))}$ for class ${i \in \{1,...,n\}}$. To make the separation between the classes in the latent space learnable, we introduce additional parameters ${\theta_\textrm{O}}$ that learn normalized center points $\mathcal{O}=\{o_1,..., o_n\} \subset \mathbb{R}^{m}$.\\
To suppress the scanner-related features, we embed the latent vectors in the latent space such that latent vectors from the same class are close to each other, and those from different classes are further apart, irrespective of the domain.
We exploit the fact that the target domain contains images of all classes of the same origin. The model learns the separation of the embeddings using data of the target domain only. 
A schematic overview in 2D for the case of $n = 2$ classes is given in Figure \ref{fig2}.
We denote the latent vector of an image of the target domain of class $i$ as $f_{i,t}$, where \textit{t} denotes the affiliation to the target domain. The center points $\mathcal{O}$ are learned considering the latent vectors  $f_{i,t}$ only, such that $o_i$ is close to $f_{i,t}$, for $i \in \{1,...,n\}$, and $o_i$ is far from $o_j$ for $i \neq j$. We force the latent vector $f_i$ of an input image $x_i$ into a hypersphere of radius $r$ centered in $o_i$.  For illustration, we use the toy example of Section \ref{problemstate}: As all elements of the target domain are blue, the two learnable center points $o_1$ and $o_2$ are separated from each other based only on the number of vertices. The color is ignored. All latent vectors of hexagons $f_1$ of the source domain should lie in a ball around $o_1$, and all latent vectors of rectangles $f_2$ should lie in a ball around $o_2$. 
\begin{figure}[h]
    \begin{center}
        \includegraphics[width=0.33\textwidth]{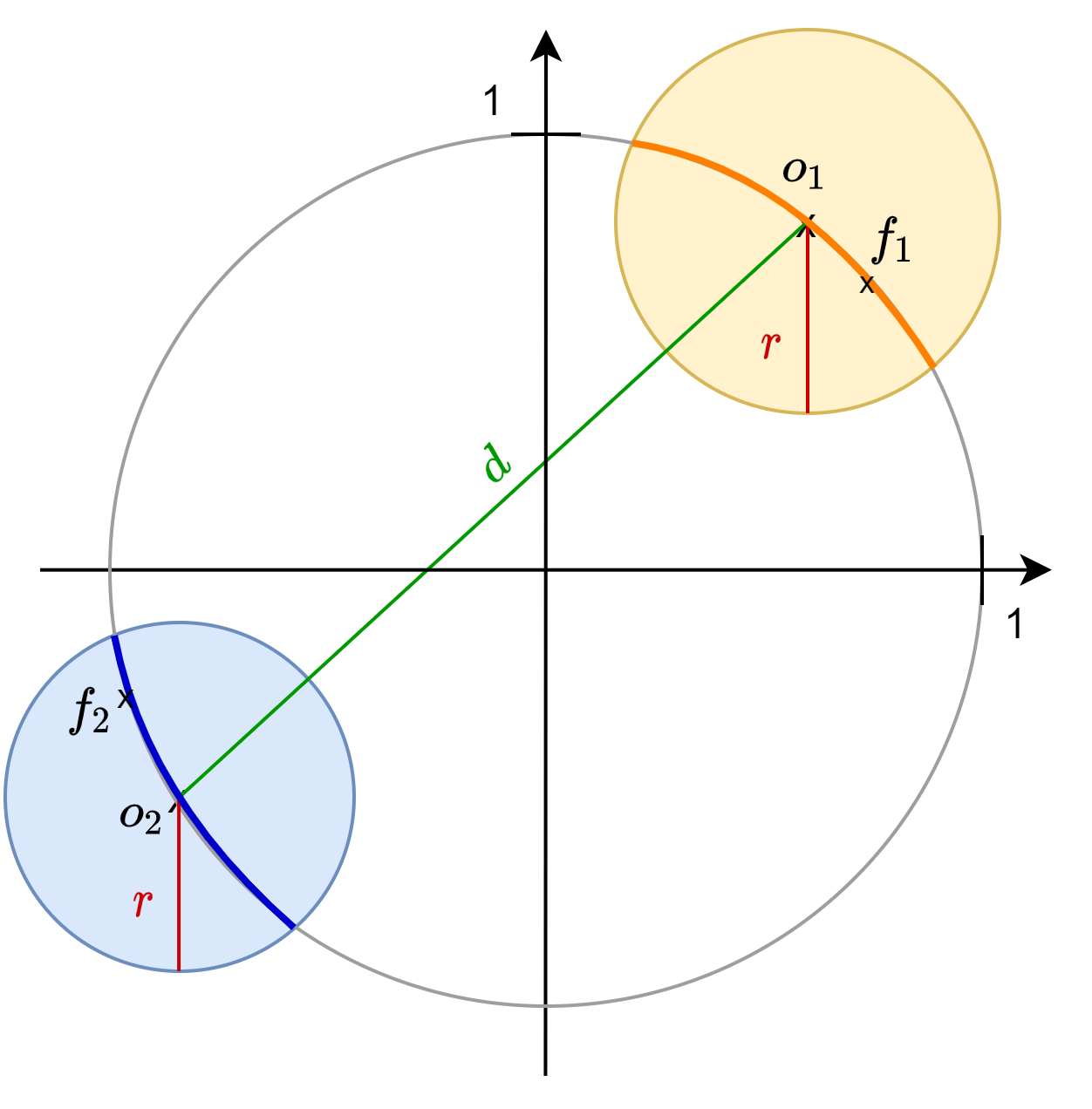}
    \end{center}
    \caption{The diagram shows a 2D sketch of the proposed latent space for $n=2$ classes, with two learnable center points $o_1$ and $o_2$. The latent vectors are normalized and lie on the unit hypersphere. Latent vectors $f_i$ of images of class $i$ should lie within the circle around $o_i$, on the orange line or blue line respectively.} \label{fig2}
\end{figure}

\subsection{Loss functions} \label{loss}
The overall objective function is given by
\begin{equation}
\mathcal{L}_\textrm{total}=\underbrace{\mathcal{L}_\textrm{cls}}_{\theta_\textrm{C}}+\underbrace{\lambda_\textrm{cen}\mathcal{L}_\textrm{cen}}_{\theta_\textrm{O}, \theta_\textrm{E}}+\underbrace{\lambda_\textrm{latent}\mathcal{L}_\textrm{latent}}_{\theta_\textrm{E}}.
\end{equation}
It consists of three components: A classification loss $\mathcal{L}_\textrm{cls}$, a center point loss $\mathcal{L}_\textrm{cen}$ for learning $\mathcal{O}=\{o_1,..., o_n\}$, and a loss $\mathcal{L}_\textrm{latent}$ on the latent space. Those components are weighted with the hyperparameters $\lambda_\textrm{latent},\lambda_\textrm{cen} \in\mathbb{R}$.
The parameters $\theta_\textrm{E}, \theta_\textrm{C}$ and $\theta_\textrm{O}$ indicate which parameters of the network are updated with which components of the loss term. While the classification loss $\mathcal{L}_\textrm{cls}$ is separate and only responsible for the final score, it is the center point loss $\mathcal{L}_\textrm{cen}$ and the latent loss  $\lambda_\textrm{latent}$ that iteratively adapt the feature space to be scanner-invariant.
With this total loss objective, any classification network can be extended by our method.

\subsubsection{Classification Loss}
The classification loss $\mathcal{L}_{\textrm{cls}, \theta_\textrm{C}}(f_i)$ is defined by the cross-entropy loss. The gradient is only calculated with respect to $\theta_\textrm{C}$, as we do not want to disturb the parameters $\theta_\textrm{E}$ with the scanner bias.

\subsubsection{Center Point Loss}
To determine the center points, we designed a novel loss function defined by the distance from a latent vector $f_{i, t}$ of the target domain to its corresponding center point $o_i$. We define a radius $r > 0$ and force the latent vectors of the target domain $f_{i,t}$ to be within a hypersphere of radius $r$ centered in $o_i$. Moreover, $o_i$ and $o_j$ should be far enough from each other for $i \neq j$. As $o_i$ is normalized to a length of one, the maximal possible distance between $o_i$ and $o_j$ equals 2.
We add a loss term forcing the distance between $o_i$ and $o_j$ to be larger than a distance $d$.
The choice of $d<2$ and $r>0$ with $d>2r$ is closely related to the choice of a margin in conventional contrastive loss terms \cite{triplet,old}. The network is not penalized for not forcing $o_i$ in the perfect position, but only to an acceptable region, such that the hyperspheres do not overlap. Then, the center point loss used to update the parameters $\theta_\textrm{O}$ and $\theta_\textrm{E}$ is given by
\begin{equation}
\mathcal{L}_{\textrm{cen}, \theta_\textrm{O}, \theta_\textrm{E}}(f_{i, t}, \mathcal{O}) =      \max(\| f_{i,t}-o_i\|_2-r,0)^2 \\
 +  \sum_{k\neq  i} \frac{1}{2} \max(d-\| o_k-o_i \|_2 ,0)^2 .    
\label{center}
\end{equation}

\subsubsection{Latent Loss}
We define the loss on the latent space, similar to the \textit{Center Loss} \cite{centerloss}, by the distance from $f_i$ to its corresponding center point $o_i$
\begin{equation}
\mathcal{L}_{\textrm{latent}, \theta_\textrm{E}}(f_i, \mathcal{O}) =  \max(\| f_i -o_i \|_2 - r,0)^2.
\label{latent}
\end{equation}
With this loss, all latent vectors $f_i$ of the training set of ${\textrm{class }i}$ are forced to be within a hypersphere of radius $r$ around the center point $o_i$. This loss is used to update the parameters $\theta_\textrm{E}$ of the encoder. By choosing $r>0$, the network is given some leeway to force $f_i$ to an acceptable region around $o_i$, denoted by the orange and blue lines in Figure \ref{fig2}.

\section{Experiments}

For the MS dataset, we collected T1-weighted images acquired with 3T MR scanners with the MPRAGE sequence from five different in-house studies.  For data privacy concerns, this patient data is not publicly available. Written informed consent was obtained from all subjects enrolled in the studies. All data were coded (i.e. pseudo-anonymized) at the time of the enrollment of the patients.
To increase the number of healthy controls, we also randomly picked MPRAGE images from the Alzheimer’s Disease Neuroimaging Initiative\footnote{Data used in preparation of this article were obtained from the Alzheimer’s Disease Neuroimaging Initiative(ADNI) database (adni.loni.usc.edu). The investigators within the ADNI contributed to the design and implementation of ADNI and/or provided data but did not participate in analysis or writing of this report.} (ADNI) dataset, the Young Adult Human Connectome Project (HCP) \cite{HCP} and the Human Connectome \mbox{Project - Aging (HCPA)} \cite{HCPA}.
More details  of the different studies are given in Table 1 of the supplementary material, including the split into training, validation, and test set. An example of the scanner bias effect for two healthy control groups of the ADNI and HCP dataset can be found in Section 3 of the supplementary material. \\
All images were preprocessed using the same pipeline consisting of skull-stripping with HD-BET \cite{hdbet}, N4 biasfield correction \cite{n4itk}, resampling to a voxel size of \mbox{1 mm$\times$1 mm$\times$1 mm}, cutting the top and lowest two percentiles of the pixel intensities, and finally an affine registration to the MNI 152 standard space\footnote{Copyright (C) 1993-2009 Louis Collins, McConnell Brain Imaging Centre, Montreal Neurological Institute, McGill University.}. All images were cropped to a size of \mbox{(124, 120, 172)}.
The dimension of the latent space is $m=128$. This results in a total number of parameters of $36,431,842$. We use the Adam optimizer \cite{adam} with ${\beta_1 = 0.9}$,  ${\beta_2 = 0.999}$, and a weight decay of  $5\cdot10^{-5}$. The learning rate for the parameters $\theta_\textrm{O}$ is ${lr_\textrm{O}=10^{-4}}$, and the learning rate for the parameters $\theta_\textrm{C}$ and $\theta_\textrm{E}$ is ${lr_\textrm{E,C}=5\cdot10^{-5}}$. 
We manually choose the hyperparameters  ${\lambda_\textrm{latent} = 1}$, ${\lambda_\textrm{cen}=100}$, $d=1.9$, and ${r=0.1}$.\\
An early stopping criterion, with a patience value of 20, based on the validation loss on the target domain, is used.
For data sampling in the training set, we use the scheme presented in Algorithm 1 in Section 2 of the supplementary material. Data augmentation includes rotation, gamma correction, flipping, scaling, and cropping.
The training was performed on an NVIDIA Quadro RTX 6000 GPU, and took about 8 hours on the MS dataset. As software framework, we used Pytorch 1.5.0.

\section{Results and Discussion} \label{sec1}
To measure the classification performance, we calculate the classification accuracy, the Cohen's kappa score \cite{cohen}, and the area under the receiver operating characteristic curve (AUROC) \cite{auroc}.
We compare our approach against the methods listed below. Implementation details and the source code of all comparing methods can be found at \url{https://gitlab.com/cian.unibas.ch/L2I}. 

\begin{itemize}
    \item Vanilla classifier (\textit{Vanilla}): Same architecture as \textit{L2I}, but only $\mathcal{L}_\textrm{cls}$ is taken to update the parameters of both the encoder $E$ and the classifier $C$.
    \item Class-aware Sampling (\textit{Class-aware}): We train the \textit{Vanilla} classifier with class-aware sampling. In every batch each class and domain is represented.
    \item  Weighted Loss (\textit{Weighted}):  We train the \textit{Vanilla} classifier, but the loss function  $\mathcal{L}_\textrm{cls}$ is weighted to compensate for class and domain imbalances. 
    \item Domain-Adversarial Neural Network (\textit{DANN}) \cite{dann}: The classifier learns both to distinguish between the domains and between the classes. It includes a gradient reversal for the domain classification.
    
    \item Unlearning Scanner Bias (\textit{Unlearning}) \cite{unlearning}: A confusion loss aims to maximally confuse a domain predictor, such that only features relevant for the main task are extracted.
    
    \item Supervised Contrastive Learning (\textit{Contrastive}) \cite{supervised}: The latent vectors are pushed into clusters far apart from each other, with a sampling scheme and a contrastive loss term allowing for multiple positives and negatives.
    
    \item Contrastive Adaptation Network (\textit{CAN}) \cite{cdd}: This  state-of-the-art DA method combines the maximum mean discrepancy of the latent vectors as loss objective, class-aware sampling, and clustering.
    
    \item Fixed Center Points (\textit{Fixed}): We train \textit{L2I}, but instead of learning the centerpoints $o_1$ and $o_2$ using the target domain, we fix the center points at\\
    $o_i=\dfrac{v_i}{\|v_i\|_2}$ for $i \in \{1,2\}$, with $v_1 = (1, ..., 1)$ and $v_2=(-1,..., -1)$ .
    
    \item No margin (\textit{No-margin}): We train \textit{L2I} with $d=2$ and $r=0$, such that no margin is chosen in the contrastive loss term in Equations \ref{center} and \ref{latent}.
\end{itemize}
We report the mean and standard deviation of the scores on the test set for 10 runs. For each run the dataset is randomly divided into training, validation, and test set. In the  first three lines of Table \ref{tab4}, the scores are shown when \textit{Vanilla},  \textit{Class-aware}, and  \textit{Weighted} are trained only on the target domain. The very poor performance is due to overfitting on such a small dataset. Therefore, the target domain needs to be supplemented with other datasets.
In the remaining lines of  \mbox{Table \ref{tab4}}, we summarize the classification results for all methods when trained on the source and the target domain. 
\begin{table}[h]
	\caption{Mean [standard deviation] of the scores on the test set for 10 runs.}    \label{tab4}
	\begin{center}
		\begin{tabular}{l l|ccc|ccc}
			\toprule
		\parbox[t]{3mm}{\multirow{2}{*}{\rotatebox[origin=c]{90}{set}}}	&& \multicolumn{3}{c|}{\textbf{Target Domain}} & \multicolumn{3}{c}{\textbf{Source Domain}} \\
		&	& \textbf{accuracy} & \textbf{kappa} &  \textbf{AUROC} & \textbf{accuracy} & \textbf{kappa} &  \textbf{AUROC}   \\
			\midrule
	 \parbox[t]{3mm}{\multirow{3}{*}{\rotatebox[origin=c]{90}{target}}}	&	Vanilla   & 50.0 [0.0] &0.0 [0.0] &59.5 [13.0] \\
		&	\quad $\cdot$ Class-aware      & 65.0 [5.0]&29.3 [9.5]& 68.6 [12.4]\\
	&		\quad $\cdot$ Weighted    & 50.3 [1.1] &0.0 [0.0] &73.3 [13.5]  \\
			\midrule
	\parbox[t]{3mm}{\multirow{10}{*}{\rotatebox[origin=c]{90}{target and source}}}		&	Vanilla  & 69.0 [6.1]& 38.0 [12.1]& 79.8 [7.3] &90.3 [5.7] &80.7 [11.3] &95.0 [5.9]\\
		&	\quad $\cdot$ Class-aware      & 71.0 [8.3]&42.0 [16.6]&79.3 [13.3]&90.7[4.2]&81.3 [8.3]& 95.3 [3.9]\\
	&		\quad $\cdot$ Weighted      & 71.3 [6.3]& 42.7 [12.7] & 81.2  [7.8] &92.2 [3.1]&84.6 [6.3]&95.1 [3.4]  \\
		&	DANN       & 67.0 [5.7]& 34.0 [11.5]&74.7 [8.4]&93.1 [2.8]&86.3 [5.5]&98.7 [0.7]\\
		&	Unlearning    & 70.7 [6.6]& 41.3 [13.3]& 81.7 [9.2] &85.5 [3.2]&71.0 [6.5]&90.5 [3.3]\\
		&	Contrastive    & 76.3 [6.7]& 52.6 [13.5]&    86.7 [9.5]  &\textbf{94.5 [2.4]}& \textbf{89.0 [4.7]} &\textbf{98.9 [0.7]}   \\
		&	CAN    &75.3 [6.9] &50.7 [13.8] &83.8 [8.8]&92.8 [2.5]&85.7 [5.0]&96.7 [1.8]\\
		&	L2I [Ours] & \textbf{89.0 [3.9]}& \textbf{78.0 [7.7]} &\textbf{89.7 [7.6]}&92.0 [2.5] & 84.0 [4.9]& 91.7 [4.9] \\
			
		&	\quad $\cdot$Fixed  & 71.7[7.2]& 43.3 [14.5]& 76.5 [11.4]&90.5 [3.9]&81.0 [7.9]&90.2 [5.8] \\
		&	\quad $\cdot$No-Margin&82.0 [3.9] &64.0 [7.8] & 75.3 9.1]&91.7 [4.8]&83.3 [9.7]&84.9 [3.9] \\
			\bottomrule
		\end{tabular}
	\end{center}
\end{table}
Our method \textit{L2I} strongly outperforms all other methods on the target domain. Although we favor the target domain during training, we see that \textit{L2I} still has a good performance on the source domain. Therefore, we claim that the model learned to distinguish the classes based on disease-related features that are present in both domains, rather than based on scanner-related features.
The benefit of learning $o_1$ and $o_2$ by taking only the target domain into account can be seen when comparing our method against \textit{Fixed}. Moreover, by comparing \textit{L2I} to \textit{No-margin}, we can see that choosing $d<2$ and $r>0$ brings an advantage.
All methods perform well on the source domain, where scanner-related features can be taken into account for classification. However, on the target domain, where only disease-related features can be used, the \textit{Vanilla}, \textit{Class-aware} and \textit{Weighted} methods show a poor performance.
A visualization of the comparison between the  \textit{Vanilla} classifier and our method \textit{L2I} can be found in the t-SNE plots in Section 4 of the supplementary material. 
\textit{DANN} and \textit{Contrastive}, as well as  the state-of-the-art methods \textit{CAN} and \textit{Unlearning} fail to show the performance they achieve in classical DA tasks. 
 Although \textit{CAN} is an unsupervised method, we think that the comparison to our supervised method is fair, since \textit{CAN} works very well on classical DA problems.

\section{Conclusion}
We presented a method that can ignore image features that are induced by different MR scanners. We designed specific constraints on the latent space and define two novel loss terms, which can be added to any classification network. The novelty lies in learning the center points in the latent space using images of the monocentric target domain only. Consequently, the separation of the latent space is learned based on task-specific features, also in cases where the main task cannot be learned from the source domain alone. Our problem therefore differs substantially from classical DA or contrastive learning problems. 
We apply our method \textit{L2I} on a classification task between multiple sclerosis patients and healthy controls on a multi-site MR dataset. 
Due to the scanner bias in the images, a vanilla classification network and its variations, as well as classical DA and contrastive learning methods, show a weak performance. 
\textit{L2I} strongly outperforms state-of-the-art methods on the target domain, without loss of performance on the source domain, improving the generalization quality of the model.
Medical images acquired with different scanners are a common scenario in long-term or multi-center studies. Our method shows a major improvement for this scenario compared to state-of-the-art methods. We plan to investigate how other tasks like image segmentation will improve by integrating our approach.

%
%
%
\bibliographystyle{splncs04}
\bibliography{neurips_2021}
\end{document}


\title{Supplementary Material}
%
%
\author{}
\authorrunning{J. Wolleb et al.}
\institute{}
\maketitle              

\section{Study details of the MS dataset}
\begin{table}
    \caption{Overview of the studies in the MS dataset.  All images are T1-weighted  MPRAGE images acquired with a 3T Siemens scanner. The dataset is split proportionally into training, validation and test set. Since there are only few images of healthy subjects in Study~1, we add 9 images of healthy subjects of Study~5. Those images were also acquired on the same scanner, but they differ in the acceleration factor from images of Study~1. To make sure that this slight difference does not disturb the training or provide biased results during test time, we use those images only for validation.}    \label{sample-table-ms}
    \begin{center}

            \begin{tabular}{l l|cc|cc|cc|l l c}
                \toprule
                & & \multicolumn{2}{c|}{\textbf{Training}} & \multicolumn{2}{c|}{\textbf{Validation}} & \multicolumn{2}{c|}{\textbf{Testing}}&\textbf{Scanner}& \textbf{Age} & \textbf{Gender[m/f]}   \\
                & & healthy& MS &  healthy& MS& healthy& MS  \\
                \midrule

               \parbox[t]{3mm}{\multirow{2}{*}{\rotatebox[origin=c]{90}{\textbf{ Target}}}}   
                &Study 1 & 16 & 55& &9& 15&15&Trio Trim &19-62&34 / 58\\

                &Study 5 &&  &9&&&& Trio Trim&26-55& 4 / 5\\
\addlinespace[6pt]
                \midrule
                    \parbox[t]{3 mm}{\multirow{6}{*}{\rotatebox[origin=c]{90}{\textbf{Source}}}}&Study 2   & & 368&    &20  &&30 & Skyra-fit, Skyra& 20-76 &137 / 279  \\
                
                &ADNI     & 113& & 6&&9&& Trio Trim & 56-78& 54 / 74\\
                &HCP     & 89& & 4&&7&& Connectome &22-35& 49 / 51\\
                &HCPA   &140& & 7&&10&&Prisma &36-88 &66 / 91\\
                &Study 3   &40& &2&&3&&Skyra&27-78&21 / 31 \\
                &Study 4  &22& &1&&1&& Skyra& 21-47&12 / 12\\
                \bottomrule
            \end{tabular}

    \end{center}

\end{table}

\section{Sampling algorithm on the MS dataset}
\begin{algorithm}[h]
    \caption{Sampling scheme}
    \label{alg:example}
    \begin{algorithmic}
        \REPEAT
        \STATE \textbullet~ Sample a random batch from the whole training set (target and source domain) of size 10. Use this batch to calculate $\mathcal{L}_\textrm{cls}$ and $\mathcal{L}_\textrm{latent}$.\\
        \STATE \textbullet~ Sample one image from the target domain of each class and calculate $\mathcal{L}_\textrm{center}$.
        \STATE \textbullet~ Update network parameters with $\mathcal{L}_\textrm{total}$.
        \UNTIL{early stopping criterion on validation loss is reached.}
    \end{algorithmic}
\end{algorithm}

\section{Histogram showing the scanner bias effect}
\begin{figure}[H]
    \centerline{\includegraphics[width=0.45\textwidth]{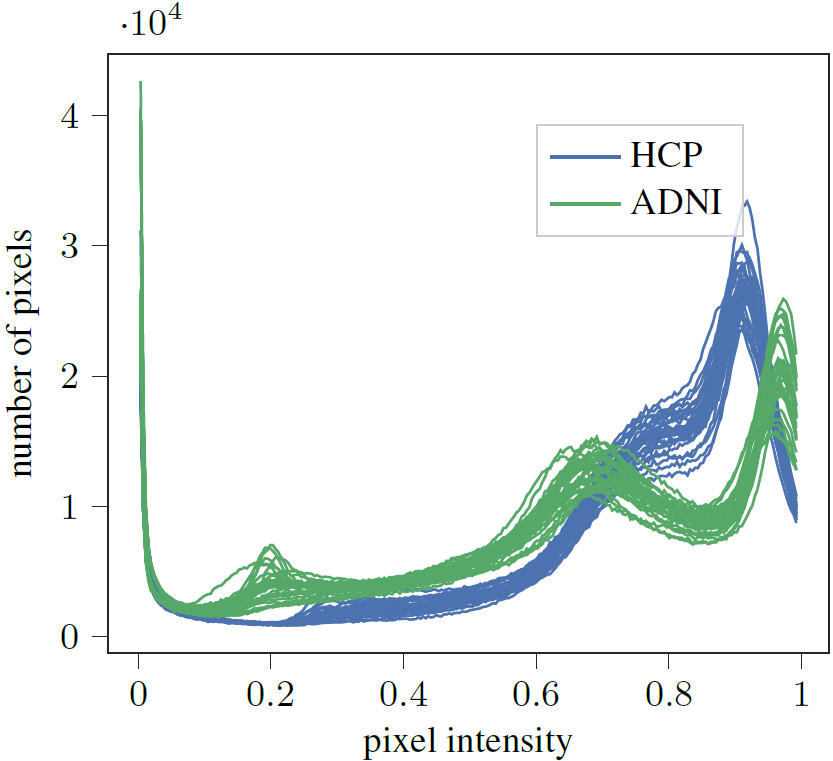}}
    \caption{Distribution of the pixel intensities for preprocessed and normalized MPRAGE MR images of healthy subjects of the ADNI and the HCP dataset. It can clearly be seen that the distribution of the pixel intensities is still different for each dataset after preprocessing. This difference originates from variations in scanner hardware, software version or scanning protocols. }\label{fig5}
\end{figure}

\section{T-SNE plots}
\begin{figure}[h]
\centering
\begin{minipage}[h]{0.43\textwidth}
\centering
 \includegraphics[width=\textwidth]{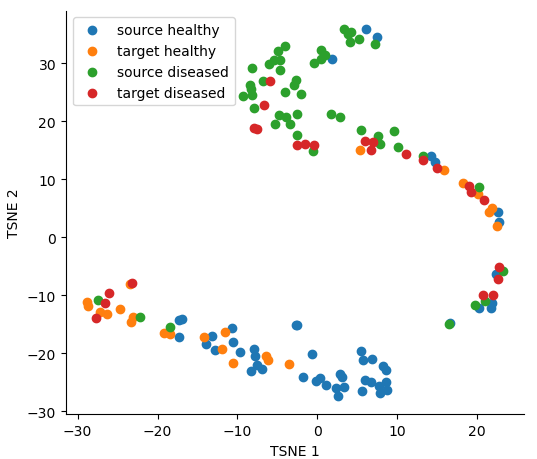}
	\end{minipage}
	\qquad \qquad
	\begin{minipage}[h]{0.43\textwidth}
	\centering
 \includegraphics[width=\textwidth]{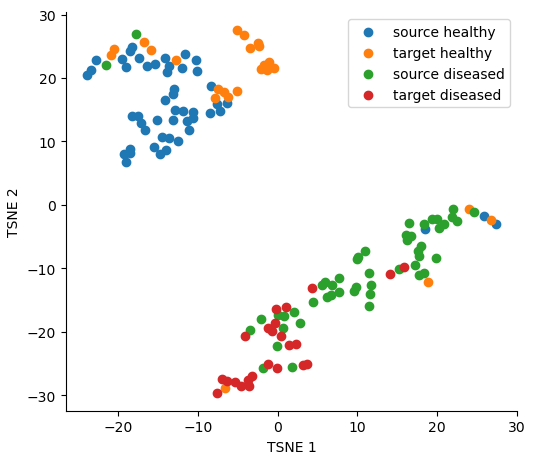}
	\end{minipage}
	\label{tsne1}
		\caption{On the left, we present the t-SNE plot of the latent vectors using the \textit{Vanilla} classifier. On the target domain (orange and red datapoints), the distinction between healthy and diseased subjects cannot clearly be seen. On the right, we show the t-SNE plot of the latent vectors using \textit{L2I}. The separation between healthy and diseased subjects can be seen, regardless of the affiliation to the source or target domain. } 
\end{figure}